\documentclass[10pt, a4paper]{article}

\usepackage{lrec2022} 

\usepackage{multibib}
\newcites{languageresource}{Language Resources}
\usepackage{graphicx}
\usepackage{tabularx}
\usepackage{soul}
\usepackage{enumitem}

\usepackage{titlesec}
\titleformat{\section}{\normalfont\large\bfseries\center}{\thesection.}{1em}{}
\titleformat{\subsection}{\normalfont\SmallTitleFont\bfseries\raggedright}{\thesubsection.}{1em}{}
\titleformat{\subsubsection}{\normalfont\normalsize\bfseries\raggedright}{\thesubsubsection.}{1em}{}
\renewcommand\thesection{\arabic{section}}
\renewcommand\thesubsection{\thesection.\arabic{subsection}}
\renewcommand\thesubsubsection{\thesubsection.\arabic{subsubsection}}

\usepackage{epstopdf}
\usepackage[utf8]{inputenc}

\usepackage{hyperref}
\usepackage{xstring}

\usepackage{color}
\usepackage{devanagari}

\title{Corpus for Automatic Structuring of Legal Documents}

\name{Prathamesh Kalamkar$^{1,2,*}$, Aman Tiwari$^{1,2,*}$, Astha Agarwal$^{1,2,*}$, Saurabh Karn$^{3,*}$\thanks{* Authors contributed equally}, \\ {\bf \large Smita Gupta$^{3}$, Vivek Raghavan$^{1}$, Ashutosh Modi$^{4}$} } 

\address{$^{1}$EkStep Foundation,  $^{2}$Thoughtworks Technologies India Pvt Ltd., \\ $^{3}$Agami , $^{4}$ Indian Institute of Technology Kanpur (IIT-K) \\ 
\{prathamk, aman.tiwari, astha.agarwal\}@thoughtworks.com,\\ \{saurabh, smita\}@agami.in, vivek@ekstep.org,  ashutoshm@cse.iitk.ac.in}

\abstract{In populous countries, pending legal cases have been growing exponentially. There is a need for developing techniques for processing and organizing legal documents. In this paper, we introduce a new corpus for structuring legal documents. In particular, we introduce a corpus of legal judgment documents in English that are segmented into topical and coherent parts. Each of these parts is annotated with a label coming from a list of pre-defined \textit{Rhetorical Roles}. We develop baseline models for automatically predicting rhetorical roles in a legal document based on the annotated corpus. Further, we show the application of rhetorical roles to improve performance on the tasks of summarization and legal judgment prediction. We release the corpus and baseline model code along with the paper.  
\\ \newline }

\begin{document}
\maketitleabstract

\section{Introduction}

In populous countries (e.g., India), pending legal cases have been growing exponentially. For example, according to India's National Judicial Data Grid, as of December 2021, there are approximately 40 million cases pending in various courts of the country \cite{njdc-district}. India follows a common-law system; consequently, due to subjectivity involved in the legal process, it may not be possible to automate the entire judicial pipeline completely; nevertheless, many intermediate tasks can be automated to augment legal practitioners, and hence expedite the system. For example, legal documents can be processed with the help of Natural Language Processing (NLP) techniques to organize and structure the data to be amenable to automatic search and retrieval. However, legal texts are different from commonly occurring texts typically used to train NLP models. Legal documents are quite long, running into tens (sometimes hundreds) of pages. Long documents make automatic processing challenging as information is spread throughout the document \cite{malik-etal-2021-ildc}. Another challenge with legal documents is the use of different lexicons. Though legal documents use natural language (e.g., English), many commonly occurring words/terms have different legal connotations. The use of different lexicons makes it challenging to adapt existing NLP models to legal texts \cite{malik-etal-2021-ildc}. Moreover, in countries like India, legal documents are manually typed and are highly unstructured and noisy (e.g., spelling and grammatical mistakes). Above mentioned challenges make it difficult to apply existing NLP models and techniques directly, which calls for the development of legal domain-specific techniques. 

\begin{figure*}[h]
\begin{center}
\includegraphics[scale=0.35]{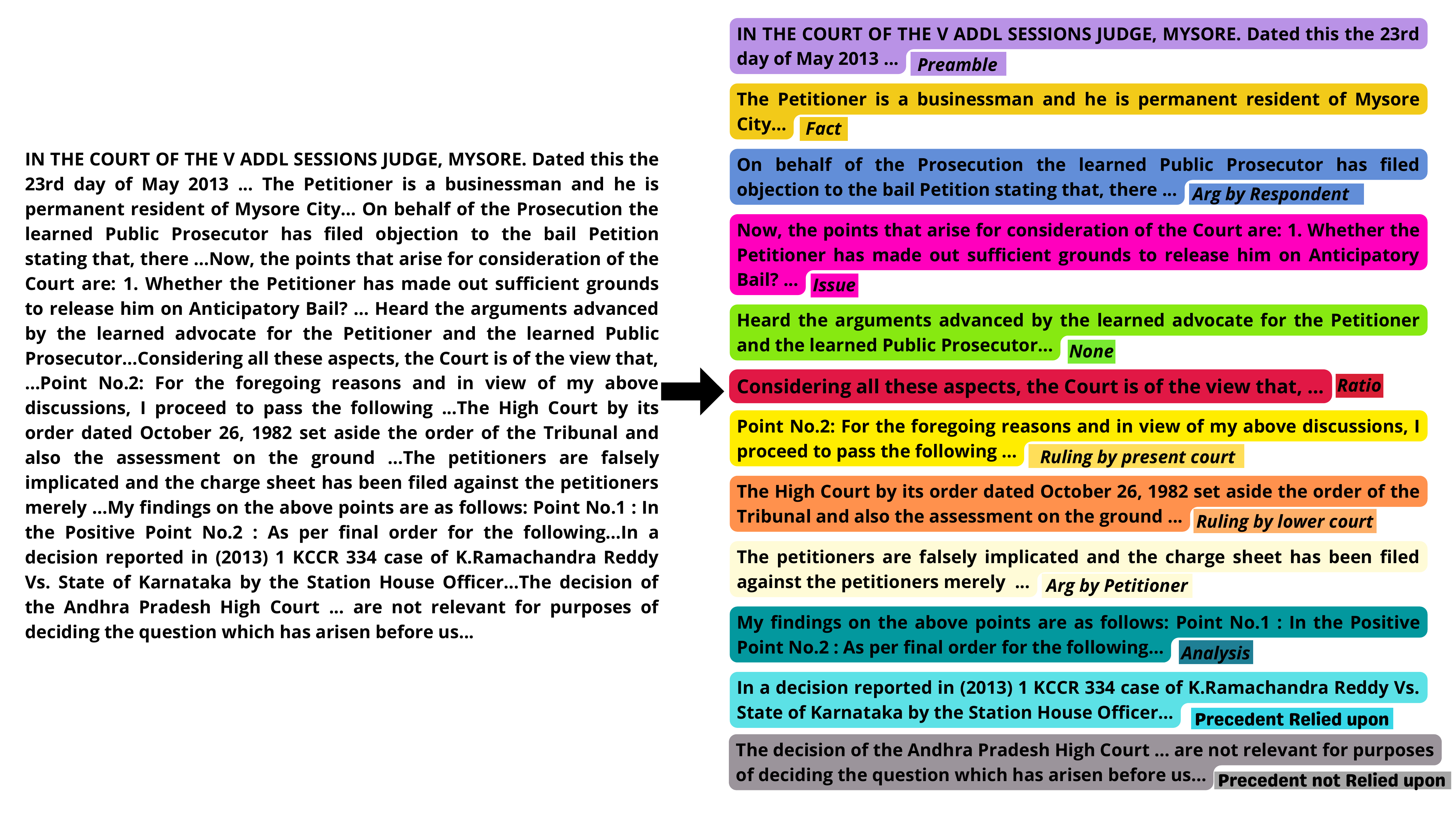} 
\caption{Example of document segmentation via Rhetorical Roles labels. On the left is excerpt from a legal document and on the right is document segmented and labelled with rhetorical role labels.}
\label{fig:example}
\end{center}
\end{figure*}

Existing state-of-the-art models in NLP are data-driven and are trained on annotated corpora. However, the legal domain suffers from the deficiency of availability of annotated corpora. It has hindered the growth of the Legal NLP domain. For example, much of the recent success in the computer vision community can be owed to the creation and availability of annotated vision corpora such as ImageNet \cite{imagenet_cvpr09,ILSVRCanalysis_ICCV2013,ILSVRC15}. In this paper, we contribute to creating annotated legal text corpora. In particular, we create a new corpus of Indian legal judgments in English that are structured and annotated with topically coherent semantic units. Since legal documents are long and unstructured, these can be divided into topically coherent parts (e.g., facts, arguments) referred to as \textit{Rhetorical Roles} \cite{saravanan2008automatic,bhattacharya2019identification,malik-rr-2021}. In this paper, with the help of legal experts, we annotate legal documents with 12 different Rhetorical Roles (RRs) (details in \S \ref{sec:rr}). An example text annotated with some of the RRs is shown in Figure \ref{fig:example}. As shown in the figure, an unstructured legal judgment document is segmented into semantically coherent parts, and each part is annotated with a rhetorical role label such as preamble, fact, ratio, etc. We experimented with different levels of granularity (phrase level, sentence level, paragraph level) for annotating RRs and decided to go for sentence-level RR annotations based on initial experiments. Each sentence in a legal document is annotated with a rhetorical role label in the proposed corpus. Typically, consecutive sentences can have a similar role in a judgment document. The rhetorical role corpus is part of a general open-source effort of creating various legal corpora for promoting the development and bench-marking of legal NLP systems. This project is called BUILDNyAI.\footnote{The word BUILDNyAI is a code-mixed (English+Hindi) term having English word BUILD and Hindi word nyAI (short for nyayi, which means justice). The project is hosted at \url{https://legal-nlp-ekstep.github.io/Competitions/Rhetorical-Role/}} We make the following contributions in this paper: 

\begin{itemize}[noitemsep,nolistsep]
\item We create a corpus of 354 Indian legal documents annotated with rhetorical roles. The corpus has 40,305 sentences annotated with 12 different RRs. To the best of our knowledge, this is the largest corpus of legal documents annotated with RRs.
\item  In order to be of practical value, using the corpus, we develop a transformer-based baseline model for automatically annotating legal documents with sentence-level RR.  
\item We show two use-cases for RRs. In particular, we show applications of RRs to the task of legal case summarization and legal judgment prediction. 
\item We release the corpus and the model implementations: \url{https://legal-nlp-ekstep.github.io/Competitions/Rhetorical-Role/}
\end{itemize}

\section{Related Work}

In recent times, there has been lot of work in the area of legal text processing. Different tasks and techniques have been proposed. For example, Prior Case Retrieval \cite{jackson2003information}, Summarization \cite{moens1999abstracting,saravanan2007using}, Case Prediction \cite{malik-etal-2021-ildc,chalkidis-etal-2019-neural,strickson-legal-2020,sulea-etal-2017-predicting,kapoor-etal-2022-HLDC}, Argument Mining \cite{wyner2010approaches,moens2007automatic}, 
Information Extraction and Retrieval \cite{tran2019building,grabmair2011toward,tran2019building}, and 
Event Extraction \cite{lagos2010event,maxwell2009evaluation,lagos2010event}. 

Recently, efforts have been made to develop corpora that could aid various legal NLP tasks; for example,  \newcite{malik-etal-2021-ildc} have released a corpus of 35K Indian Supreme Court documents for the task of judgment prediction and explanation.  \newcite{chalkidis-etal-2019-neural} have released 11,478 legal documents corresponding to the European Court of Human Rights (ECHR). \newcite{strickson_legal_2020} have proposed a corpus of 4,959 UK Supreme Court documents. \newcite{xiao2018cail2018} have created a large-scale corpus of 2.68 million criminal case documents and released CAIL (Chinese AI and Law Challenge) dataset for judgment prediction. A new multilingual dataset of European Union (EU) legal documents has been recently released by \newcite{chalkidis-etal-2021-multieurlex}.

Research in rhetorical roles for legal text processing has been active in the past few years. \newcite{Farzindar2004LetSumAA,hachey2006extractive} have leveraged rhetorical roles to create summaries of legal texts. \newcite{saravanan2008automatic} proposed a CRF-based model using hand-crafted features for segmenting documents using seven different roles. \newcite{bhatia2014analysing} created Genre Analysis of Legal Texts to create seven rhetorical categories. \newcite{bhattacharya2019identification} have proposed CRF-BiLSTM model for automatically assigning rhetorical roles to sentences in Indian legal documents. \cite{malik-rr-2021} have created a RR corpus and annotated with 13 fine-grained roles and further they have developed a multi-task learning based model for predicting RR. In this paper, we also propose a corpus of English Indian legal judgment documents annotated with Rhetorical Roles; however, we annotate the documents with a more extensive set of 12 rhetorical role labels and a NONE label (in the case none of the 12 labels are applicable). Moreover, to the best of our knowledge, we create the largest corpus of 354 documents (vs. 100 documents in previous RR corpus by \newcite{malik-rr-2021}), with 40,315 sentences annotated with 13 (12 + NONE) different types of rhetorical role labels. We propose state-of-the-art transformer models for RR prediction and show the use case of RRs for case summarization and legal judgment prediction.    

Recent success in almost every area in NLP has been due to transformer-based neural architectures \cite{wang2018glue}. We do not discuss the details of transformer architectures here and refer the reader to the survey on transformers by \newcite{tay2020efficient}. We develop transformer-based baseline models for automatically segmenting legal documents into RRs units. 

\section{Rhetorical Roles Corpus} \label{sec:rr}

As outlined earlier, legal documents are typically long, and information is spread throughout the document. In order to make the automatic processing of documents easier, documents are divided into topically coherent segments referred to as Rhetorical Roles \cite{malik-rr-2021}. In this paper, we propose the use of 12 RRs and a NONE label. We started with the list of RR labels proposed by \newcite{bhattacharya2019identification}; however, we found some of the RR to be ambiguous, hence after having elaborate discussions with law professors, we split some of the RRs (arguments and precedents) to arrive at the list of 12 main roles. Details and definitions for each of the RR are as follows:

\begin{itemize}[noitemsep,nolistsep]
\item \textbf{Preamble (PREAMBLE):} This covers the metadata related to the legal judgment document. A typical judgment would start with the court name, the details of parties, lawyers and judges' names, headnote (summary). This section typically would end with a keyword like (JUDGMENT or ORDER). Some documents also have HEADNOTES, ACTS sections in the beginning. These are also part of the Preamble.

\item \textbf{Facts (FAC):} This corresponds to the facts of the case. It refers to the chronology of events that led to filing the case and how it evolved (e.g., First Information Report (FIR) at a police station, filing an appeal to the Magistrate, etc.) Depositions and proceedings of the current court, and summary of lower court proceedings. 

\item \textbf{Ruling by Lower Court (RLC):} Cases are not directly filed in the higher courts but are appealed from lower courts.  Consequently, the documents contain judgments given by the lower courts (Trial Court, High Court) based on the present appeal (to the Supreme Court or high court). The lower court's verdict, analysis, and the ratio behind the judgment by the lower court is annotated with this label.

\item \textbf{Issues (ISSUE):} Some judgments mention the key points on which the verdict needs to be delivered. Such Legal Questions Framed by the Court are ISSUES. 

\item \textbf{Argument by Petitioner (ARG\_PETITIONER):} Arguments by petitioners' lawyers. Precedent cases argued by petitioner lawyers fall under this category, but when the court discusses them later, then they belong to either the relied / not relied upon category.

\item \textbf{Argument by Respondent (ARG\_RESPONDENT):} Arguments by respondents' lawyers. Precedent cases argued by respondent lawyers fall under this, but when the court discusses them later, they belong to either the relied / not relied category. 

\item \textbf{Analysis (ANALYSIS):} These are views of the court. This includes courts' discussion on the evidence, facts presented, prior cases, and statutes. Discussions on how the law is applicable or not applicable to the current case. Observations (non-binding) from the court. It is the parent tag for three tags: PRE\_RLEIED, PRE\_NOT\_RELIED, and STATUTE i.e., every statement which belongs to these three tags should also be marked as ANALYSIS. 

\item \textbf{Statute (STA):} This includes texts in which the court discusses established laws, that can come from a mixture of sources: Acts , Sections, Articles, Rules, Order, Notices, Notifications, and Quotations directly from the bare act. The statute will have both the tags Analysis + Statute. 

\item \textbf{Precedent Relied (PRE\_RELIED):} Texts in which the court discusses prior case documents, discussions and decisions which were relied upon by the court for final decisions. Precedent will have both the tags Analysis + Precedent. 

\item \textbf{Precedent Not Relied (PRE\_NOT\_RELIED):} Texts in which the court discusses prior case documents, discussions and decisions which were not relied upon by the court for final decisions. It could be due to the fact that the situation, in that case, is not relevant to the current case.

\item \textbf{Ratio of the decision (Ratio):} This includes the main reason given for the application of any legal principle to the legal issue. It is the result of the analysis by the court. It typically appears right before the final decision. It is not the same as "Ratio Decidendi" taught in the legal academic curriculum. 

\item \textbf{Ruling by Present Court (RPC):}  Final decision +  conclusion +  order of the Court following from the natural/logical outcome of the rationale.

\item \textbf{NONE:} If a sentence does not belong to any of the above categories, it is labeled as NONE. 

\end{itemize}


\subsection{Corpus Documents}

The corpus consists of legal judgment documents from the Supreme Court of India, High Courts in different Indian states, and some district-level courts. Raw judgment text files were scraped from Indian Court websites.\footnote{\url{https://main.sci.gov.in/};  \url{https://ecourts.gov.in/ecourts_home/static/highcourts.php}} Data has a mix of Supreme Court judgments (40\%) , High Courts judgments (40\%) and district court judgments (20\%). To develop baseline models, we divided the dataset into train, and validation.  Test set was further divided into in-domain and out of domain. Train, validation and test (in-domain) datasets contain annotated judgments belonging to tax and criminal cases. Test (out-domains) contains annotated judgements from 3 domains : Motor Vehicles Act (9) , Industrial and Labour law (8) and Land and Property law (10). The statistics of the corpus are shown in Table \ref{table:Datasets summary statistics}. Table \ref{table:roleNumbers} gives number of sentences for each role in the entire corpus. Qualified law experts annotated test data with cross checks. 

\begin{table}[t]
\small
\begin{center}
\begin{tabularx}{\columnwidth}{|l|X|X|X|X|}
    \hline
        \textbf{Dataset} & \textbf{Docs} & \textbf{Sen- tences} & \textbf{Tokens} & \textbf{Avg Tokens} \\ \hline
        Train & 247 & 28986 & 938K & 3797 \\ \hline
        Validation & 30 & 2879 & 88K & 2947 \\ \hline
        Test (in-domain) & 50 & 4158 & 134K & 2681 \\ \hline
        Test (out-domain) & 27 & 4292 & 127K & 4722\\\hline
        \textbf{Total} & \textbf{354} & \textbf{40315} & \textbf{1.3M} & \textbf{3638} \\ \hline
    \end{tabularx}
\caption{Corpus Statistics: The corpus is split into train, val and test. The table shows number of documents, sentences, tokens and average number of tokens per document.}
\label{table:Datasets summary statistics}
 \end{center}
 \vspace{-5mm}
\end{table}

\begin{table}[ht]
\small
\begin{center}
\begin{tabularx}{0.65\columnwidth}{|l|X|}
\hline
\textbf{Rhetorical Role} & \textbf{Sentences}  \\ \hline
ANALYSIS                 & 14300\\ \hline
ARG PETITIONER          & 1771\\ \hline
ARG RESPONDENT          & 1068\\ \hline
FAC                      & 8045\\ \hline
ISSUE                    & 535\\ \hline
NONE                     & 2037\\ \hline
PREAMBLE                 & 6116\\ \hline
PRE NOT RELIED         & 217\\ \hline
PRE RELIED              & 1934\\ \hline
RATIO                    & 1014\\ \hline
RLC                      & 1081\\ \hline
RPC                      & 1562\\ \hline
STA                      & 625\\ \hline  \hline
Overall                      & 40305\\ \hline
\end{tabularx}
 \caption{Role-wise sentence count in the entire corpus}
 \label{table:roleNumbers}
 \end{center}
  \vspace{-5mm}
 \end{table}
 

\subsection{Annotation Process}

The annotation process was designed in consultation with legal experts (law professors and legal practitioners). Given the nature of the task, the RR annotations require a deep understanding of the law and the legal process. Consequently, we involved law students and legal practitioners in annotating the documents. The process involved annotating each sentence in a given document with one of the 12 RR + None labels described earlier. We experimented with different levels of granularity (phrase level, sentence level, paragraph level, etc.) for annotating the documents with RR. Pilot experiments indicated sentence level RR annotation to be appropriate as it maintains the balance (with regard to semantic coherence) between too short and too long texts. The legal documents were split using spaCy library \cite{spacy2021}. Rhetorical role annotation is not a trivial task; we faced two main challenges in the annotation activity: availability for a large group of legal experts and, secondly, motivating the legal experts to perform annotation consistently while maintaining quality. We performed the annotation activity via crowdsourcing as described next.      

\subsection{Data Annotation Pipeline}
\begin{figure}[!h]
\begin{center}
\includegraphics[scale=0.6]{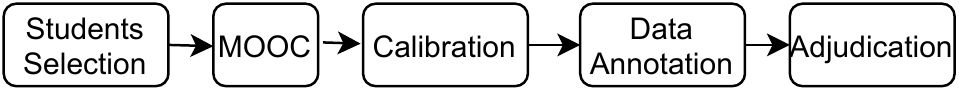} 

\caption{Data Annotation Pipeline}
\label{fig:DataAnnotationPipeline}
\end{center}
 \vspace{-5mm}
\end{figure}

Corpus documents were annotated via a crowdsourcing activity. We invited law students from various law schools across the country to volunteer for the data annotation exercise. We created processes to onboard student volunteers and introduced them to the entire activity and its goal. Filtering was carried out at multiple stages to retain the most motivated and consistent (from the perspective of quality of the annotations) students. The entire pipeline is shown in Figure \ref{fig:DataAnnotationPipeline}. We describe each stage of the pipeline in the next sections. 

\subsubsection{Student Selection} \label{sec:studentSelection}
\begin{figure}[!h]
\begin{center}
\includegraphics[scale=0.15]{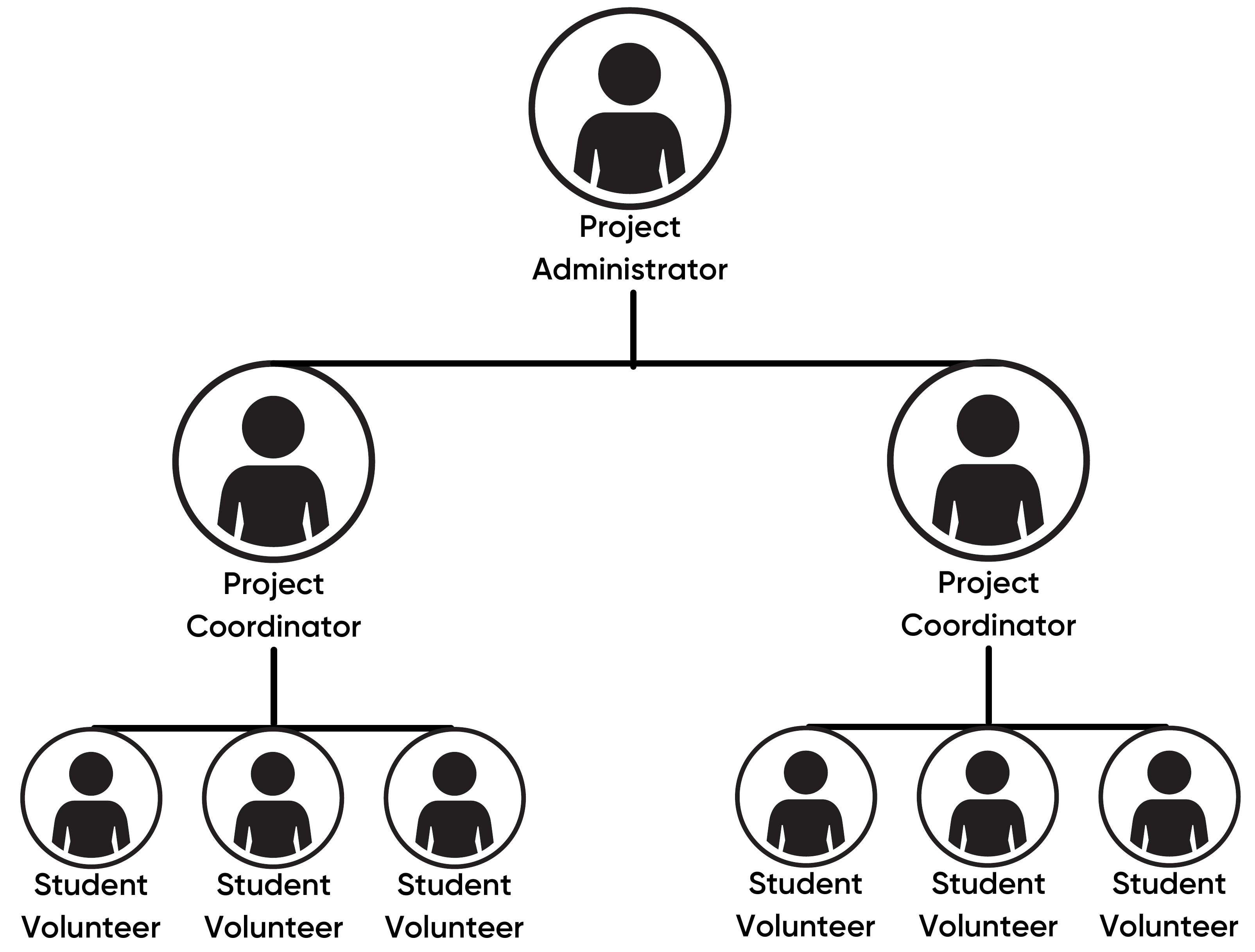} 
\caption{Organization Structure}
\label{fig:OrganizationStructure}
\end{center}
 \vspace{-5mm}
\end{figure}

We did a nationwide call for volunteers through a network of law students. The application required students to describe their motivation. A basic screening was done to eliminate applications that were partially filled. Finally, after filtering, we selected an initial group of 50 students. The selected students were then on-boarded and were motivated by explaining the big picture of the impact of their contribution. The data annotations were done voluntarily by law students from multiple Indian law universities. Interaction with the law students revealed that they were motivated to learn more about AI and contribute towards the development of the AI field, and hence they volunteered for the activity. In order to smoothly conduct the annotation activity via crowdsourcing, we organized the volunteers in a hierarchical structure based on their experience and performance during a pilot study. The organizational structure for this exercise is shown in Figure \ref{fig:OrganizationStructure}.

\noindent\textbf{Project Administrators:} They designed data collection and communication processes, built tools for data collection, and supervised the overall activity. This group included law experts and authors of the paper. 

\noindent\textbf{Project Coordinators:} They mentored and resolved the doubts of the students. They were responsible for assuring the quality of the data. Coordinators identified and rectified conceptual errors among the students. Further, the coordinators assisted the administrators during the adjudication process. 

\noindent\textbf{Student Volunteers:} They annotated the data and also provided feedback on the entire process. Volunteers were in constant communication with the coordinators. At later stages of annotations, some of the best-performing students assisted in the adjudication process (\S \ref{sec:adjudication}). Best performing students were selected based on two criteria: timely submissions and ground truth agreement score. Students were assessed if they completed the task within a stipulated time at each annotation stage. Furthermore, each batch of annotation document consisted of sentences for which true (gold) RR labels were known apriori (also \S \ref{sec:dataAnnotation}). Students were assessed for their performance on the ground truth (sentences with gold RR labels), and students who were correct on at least 90\% of ground truth sentences were considered for the best performing category.

Before beginning the entire activity, we conducted a small pilot to assess the feasibility of crowdsourcing with student volunteers. Volunteers who completed MOOC, calibration and annotation exercises with satisfactory performance were then invited to become project coordinators for the subsequent data collection phase. The chance to become coordinator further provided positive reinforcement for the efforts, thus keeping the students well motivated. In the end, we selected eight students as project coordinators. 

\subsubsection{MOOC}
Law students do not have an understanding of the workings of AI. We designed a MOOC (Massive Open Online Course)\footnote{\url{https://www.youtube.com/playlist?list=PL1z52lLL6eWnDnc3Wgfcu6neczrU3fFw0}} for the annotators. The MOOC explained the AI technologies to the law students, described the process of building datasets for AI algorithms, and explained the concept of the rhetorical role. Students were expected to complete the MOOC in a stipulated amount of time and complete the associated quiz, which checked for a basic understanding of the rhetorical role definitions. 

\subsubsection{Calibration}
Since in the initial stages, students can differ in understanding RRs. We calibrated the students to bring them to a common ground. Calibration focused on shaping a common understanding of definitions among students. Students were asked to annotate three judgments that experts had already annotated. The sentences that differed from expert (gold) annotations were highlighted, and students were asked to calibrate their annotations. Calibration was an iterative process, and it was carried out till students came at the level of expert annotations. 

\begin{figure}[t]
\begin{center}
\includegraphics[scale=0.4]{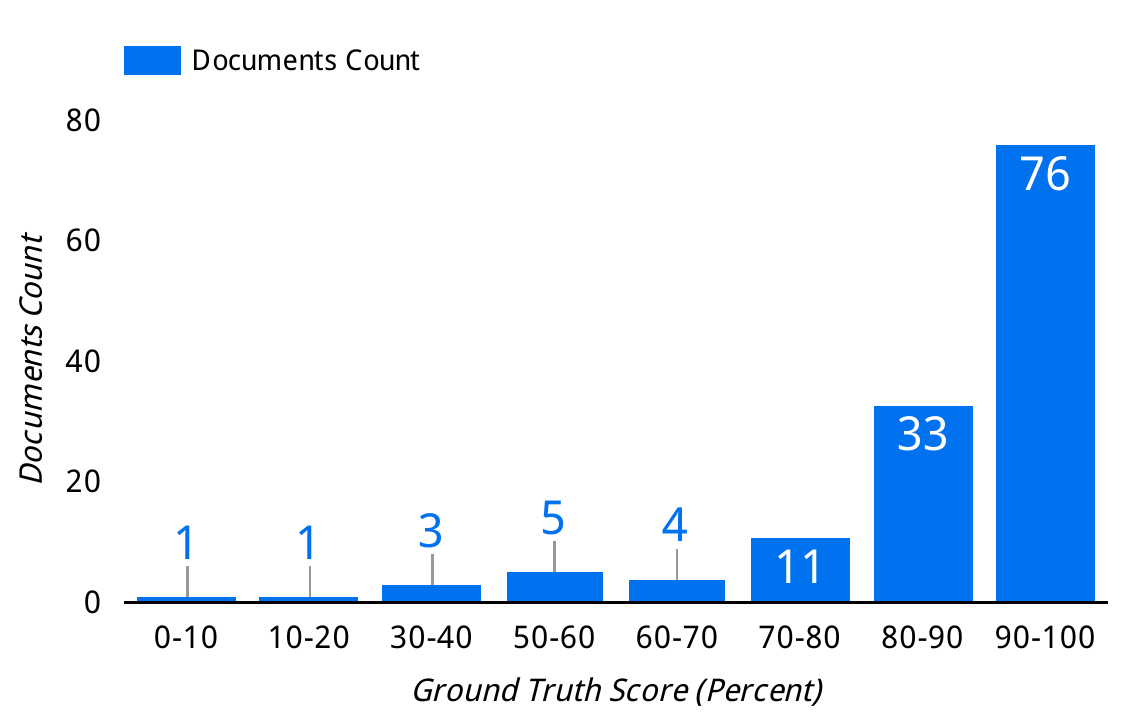}
\caption{Ground Truth Score Histogram}
\label{fig:groundTruth}
\end{center}
 \vspace{-6mm}
\end{figure}

\subsubsection{Data Annotation} \label{sec:dataAnnotation}
In the end, 35 out of 50 selected students qualified for the calibration stage, and this was the final pool that annotated the entire corpus. Each student annotated 24 documents, and three students annotated each document. We did not observe any student dropout after the calibration stage. On average, it took about 40 minutes to annotate a single document. The entire annotation activity took around six weeks. Students annotated train and validation documents ( = 277), and experts annotated 77 test documents. As described earlier, during the annotation process, each student was also randomly assigned four documents (chosen randomly with replacement from the test set) for which gold (ground truth) annotations were known to coordinators and administrators but not to the students. The performance of students (referred to as Ground Truth Score) on these gold documents was assessed. Ground truth score is the percentage of sentences in gold documents that are correctly annotated. The average ground truth score for all students was 85\%. Figure \ref{fig:groundTruth} shows histogram of ground truth scores for a judgment. It shows that the majority of documents are in the 90 to 100 percent range, indicative of consistent annotations with ground truth docs. Note that documents shown in Figure \ref{fig:groundTruth} (y-axis) are chosen randomly (with replacement) from the test set and hence there is overlap between documents across different batches. Furthermore, coordinators provided feedback to students with lower scores to improve their overall annotation quality.  

\subsubsection{Adjudication}\label{sec:adjudication}
A majority voting scheme was used to decide the final RR label. However, in some instances, annotators assigned three different labels; such documents were further sent for adjudication. The adjudication was done by experts, project coordinators, and some of the best-performing students (\S \ref{sec:studentSelection}). 

\subsubsection{Annotation Quality Assessment}
Final annotation quality was evaluated using Fleiss Kappa \cite{fleiss2013statistical}. Overall, Fleiss Kappa score was 0.59, pointing towards moderate agreement. We saw high agreement amongst annotators on PREAMBLE, RPC, NONE, and ISSUE. There were medium agreements on FACTS, RLC, ANALYSIS, PRECEDENT, and ARGUMENTS. RATIO was the most ambiguous role. ANALYSIS was very often confused with FACTS and ARGUMENTS. In a judgment, a judge emphasizes some of the facts, which as per definition, are considered as analysis role; however, annotators often confuse them as facts role. Moreover, sometimes the judge may mention arguments and give their opinion on it; this, as per definition, is the analysis role, but annotators sometimes confuse it with the argument role. FACTS was sometimes confused with RLC (Ruling by Lower Court). 

\section{RR Prediction Baseline Models}

\begin{figure}[t]
\begin{center}
\includegraphics[scale=0.35]{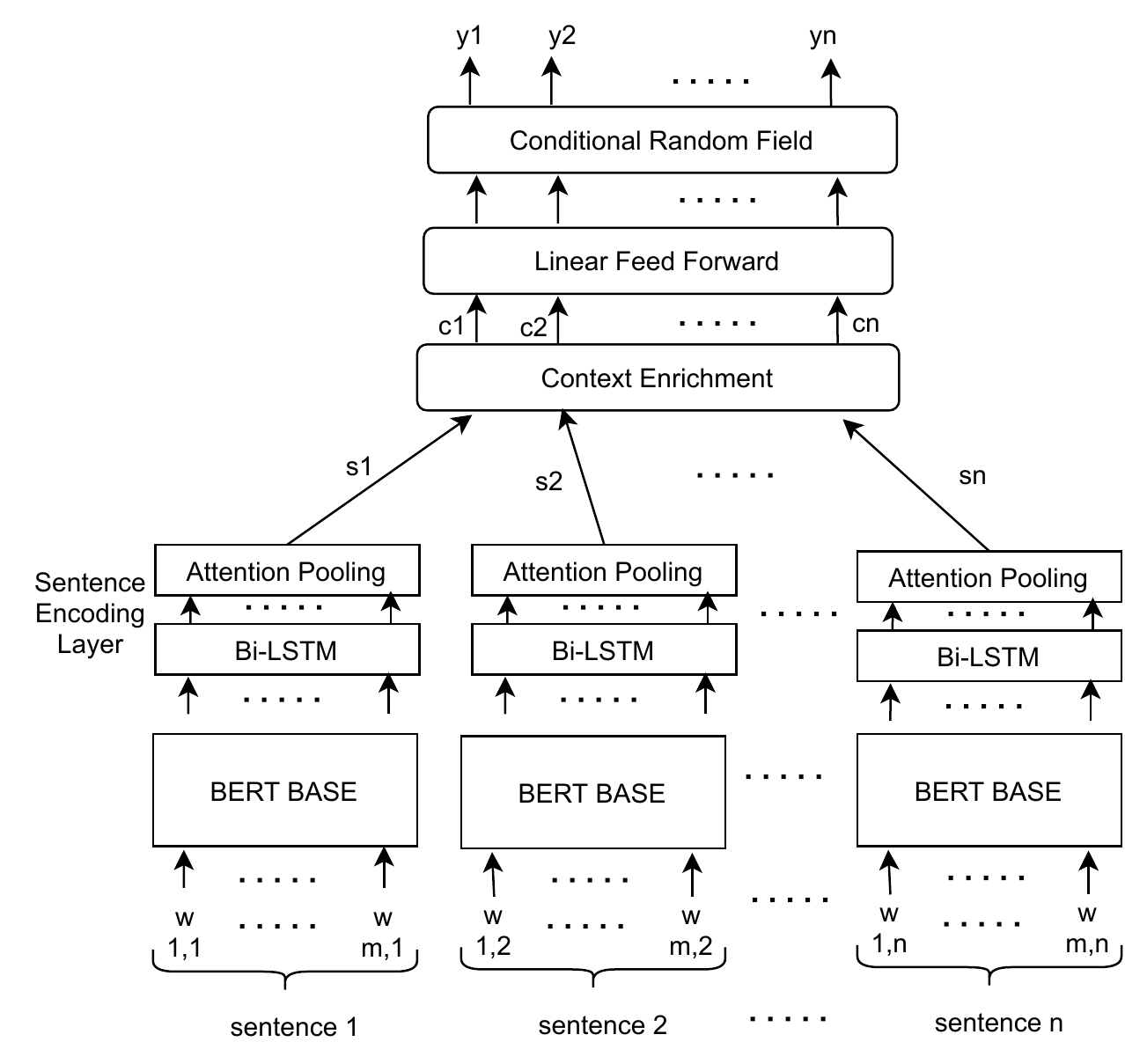}
\caption{RR Prediction Baseline model inspired by~\protect\newcite{brack2021sequential}}
\label{fig:BaselineModel}
\end{center}
\end{figure}

The end goal behind this work has been to encourage the development of systems that can segment a new legal document automatically in terms of rhetorical roles. Towards this goal, we experimented with some baseline models. Since transformer-based models \cite{wolf-etal-2020-transformers} have shown state-of-the-art (SOTA) performance on most of the NLP tasks, including the tasks in legal NLP domain \cite{malik-etal-2021-ildc}, we mainly experimented with them. In the RR prediction task, given a legal document, the task is to predict the RR label for each sentence in the document. We pose this as a multi-class sequence prediction problem. We initially experimented with variants of the model by \newcite{bhattacharya2019identification}. In particular, we use a CRF (Conditional Random Field) model for RR prediction. The features for this CRF model come from a transformer, i.e., the BERT-BASE \cite{DBLP:journals/corr/abs-1810-04805} model is used to get sentence embeddings corresponding to the CLS token. These sentence embeddings are then passed through the CRF layer to get final predictions. We call this model BERT\_CRF. We also tried the architecture proposed by \newcite{cohan-2019} which captures contextual dependencies using only BERT without the need for hierarchical encoding using a CRF. We call this model BERT\_only. After experiments with vanilla transformer models, we finally created the baseline system using the SciBERT-HSLN architecture \cite{brack2021sequential}. Figure \ref{fig:BaselineModel} shows the overall architecture of the proposed model. In the proposed model, each sentence is passed through BERT BASE model to get word embeddings, these embeddings are further processed by Bi-LSTM layer followed by attention-based pooling layer to get sentence representations $\{s_1,s_2,..s_n\}$. Context Enrichment layer encodes the contextual information, by taking sequence of sentence representations, resulting in contextualized sentence representations: $\{c_1,c_2,..,c_n\}$. This is followed by MLP layers and CRF that leverage the distributed representation features to predict the RR label for each sentence via softmax activation. 

\begin{table}[t]
\small
\begin{center}
\begin{tabularx}{0.85\columnwidth}{|X|c|c|c|}
\hline
\textbf{Model} & \textbf{Precision} & \textbf{Recall} & \textbf{F1}\\ \hline
        BERT\_CRF & 0.24 & 0.24 & 0.23\\ \hline
        BERT\_only & 0.67 & 0.68 & 0.67\\ \hline
        SciBERT-HSLN & 0.79 & 0.80 & 0.79\\ \hline

\end{tabularx}
 \caption{Performance of models on test (in-domain) data}
 \label{Comparison of Rhetorical Roles Prediction models}
 \end{center}
  \vspace{-5mm}
 \end{table}
 
\begin{figure}[t]
\begin{center}
\includegraphics[scale=0.45]{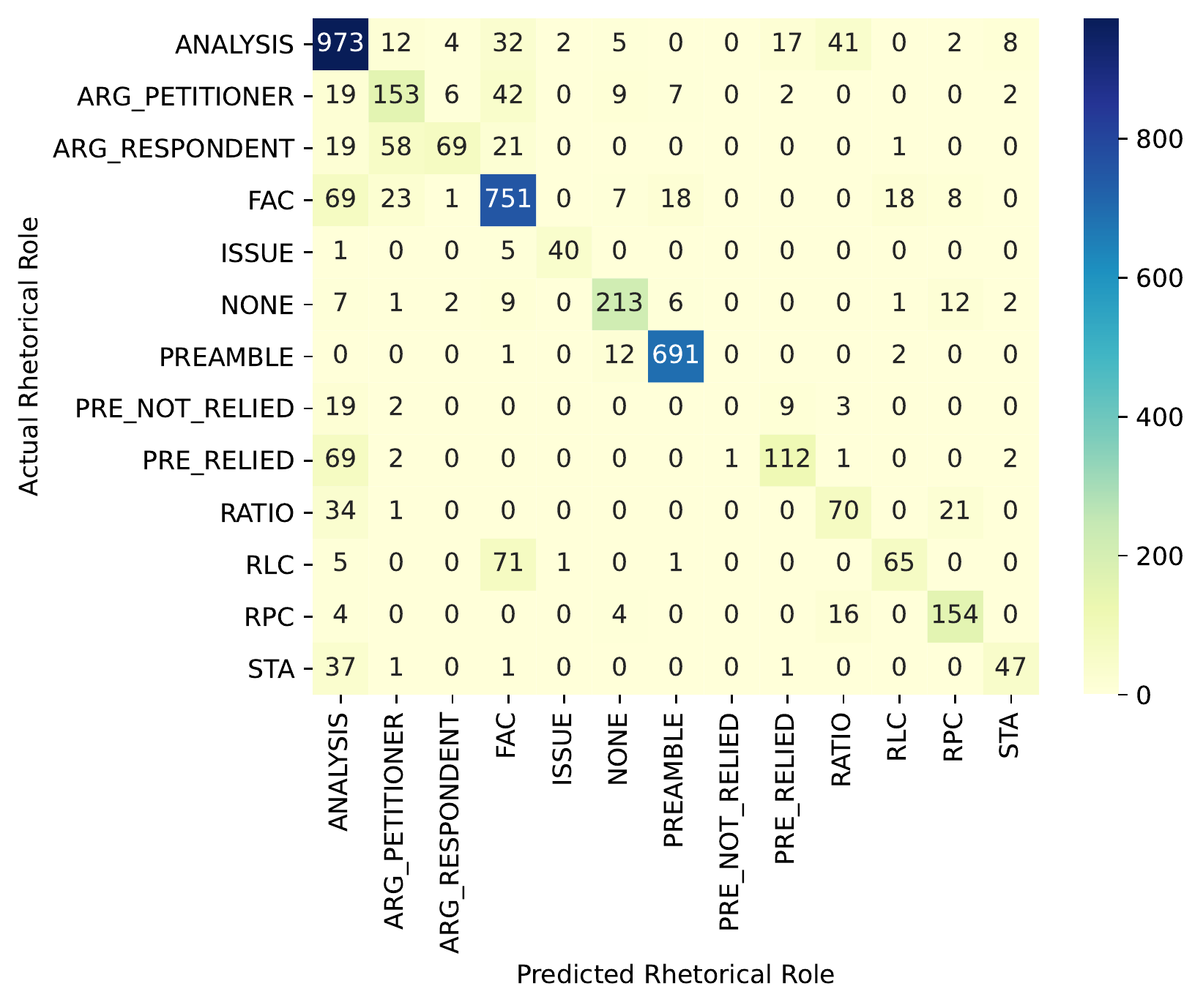} 
\caption{Confusion Matrix for SciBERT-HSLN model predictions on the test data}
\label{fig:confusion_matrix}
\end{center}
 \vspace{-5mm}
\end{figure}

\textbf{Results:} The performance of different models was tested on test(in-domain) data and results are given in Table  \ref{Comparison of Rhetorical Roles Prediction models}. We use standard weighted F1 score metric for evaluation. As can be observed, the BERT\_CRF model performs the worst, and the BERT\_only model performs worse than the proposed model SciBERT-HSLN, which achieved a weighted F1 score of 78\%. It is perhaps because SciBERT-HSLN, being a sequential model, can capture longer range dependencies between sentences in a document. The results of the model on the test set for each of the RR labels are shown in Table \ref{table:modelScores}. Figure \ref{fig:confusion_matrix} shows the confusion matrix for the SciBERT-HSLN model. As can be observed from Table \ref{table:modelScores} and Figure \ref{fig:confusion_matrix}, ARGUMENTS based roles are miss-classified very often and confused among the two types of  ARGUMENTS and also sometimes confused with FACTS and ANALYSIS. PREAMBLE is almost perfectly classified. As can be seen, PRECEDENT NOT RELIED is completely miss-classified and confused with PRECEDENT RELIED and ANALYSIS. RATIO is often confused with ANALYSIS, and this trend is similar to what was observed for annotators as well. Similar to what was observed for annotators, RPC, PREAMBLE, NONE and ISSUE are classified with decent F1 scores. STATUES are also not well classified as many times a judge mentions some laws in their opinion and model tends to learn these spurious patterns as analysis and miss-classifies actual statues as analysis. We have also created a leaderboard\footnote{\url{https://legal-nlp-ekstep.github.io/Competitions/Rhetorical-Role/}} for the task of RR prediction where other researchers can experiment with various approaches. 

\textbf{Results on test (out-domain) data:} 
In order to check if the baseline model trained on Criminal and Tax cases generalized to other domains, we tested the baseline model on 27 judgments from Motor Vehicles, Industrial and Labour and Land and Property cases. Weighted F1 reduced to 0.70. This degradation in performance is mainly due to different style of writing in the judgments. 

\begin{table}[t]
\small
\begin{center}
\begin{tabularx}{0.95\columnwidth}{|X|c|c|c|}
\hline
\textbf{Rhetorical Role} & \textbf{Precision} & \textbf{Recall} & \textbf{F1}\\ \hline
        ANALYSIS & 0.77 & 0.89 & 0.83 \\ \hline
        ARG\_PETITIONER & 0.60 & 0.64 & 0.62 \\ \hline
        ARG\_RESPONDENT & 0.84 & 0.41 & 0.55 \\ \hline
        FAC & 0.80 & 0.84 & 0.82 \\ \hline
        ISSUE & 0.93 & 0.87 & 0.90 \\ \hline
        NONE & 0.85 & 0.84 & 0.85 \\ \hline
        PREAMBLE & 0.96 & 0.98 & 0.97 \\ \hline
        PRE\_NOT\_RELIED & 0.00 & 0.00 & 0.00 \\ \hline
        PRE\_RELIED & 0.79 & 0.60 & 0.68 \\ \hline
        RATIO & 0.53 & 0.56 & 0.54 \\ \hline
        RLC & 0.75 & 0.45 & 0.57 \\ \hline
        RPC & 0.78 & 0.87 & 0.82 \\ \hline
        STA & 0.77 & 0.54 & 0.64 \\ \hline
        \hline
        Overall  & 0.79 & 0.80 & 0.79\\ \hline
\end{tabularx}
 \caption{F1 scores of RR baseline model for each of the rhetorical role on test data}
 \label{table:modelScores}
 \end{center}
 \end{table}

\section{Applications of Rhetorical Roles Prediction Task}

The purpose of creating a rhetorical role corpus is to enable automated understanding of legal documents by segmenting them into topically coherent units. This can be helpful in various applications such legal document summarization \cite{bhattacharya2019comparative}, and legal judgment prediction \cite{malik-etal-2021-ildc}. In this paper, we explore both the use-cases. We experimented with how rhetorical roles prediction could help create abstractive, extractive summaries of Indian court judgments and predict the judgment outcome based on the judgment text.

\subsection{Extractive Summarization of Court Judgments using Rhetorical Roles}
\label{subsection:extractive_summarization}

We explored the task of extractive summarization. For a given legal document, the task requires extracting the salient sentences that would summarize the document. We experimented with the LawBriefs corpus consisting of 285 extractive summaries of Indian court judgments prepared by law students from a National Law University in India. The corpus was created by providing judgment documents to law students, followed by a questionnaire that required them to pick salient sentences that would answer the questions and, in the process, create the summaries. The questions pertained to facts, arguments, issues, ratio, and decisions. We wanted to experiment with how rhetorical roles could be helpful in extracting summaries.   

We finetuned BERTSUM \cite{liu2019text} model on the Lawbriefs data to pick up the top 20\% of the sentences as summaries. Since the judgments are much longer than 512 token limits of BERTSUM, we created non-overlapping chunks of 512 tokens and created 3151 chunks in training data from 235 judgments and 827 chunks from 50 judgments as test data. We then trained another model, which also takes as input a rhetorical role for each sentence. We concatenated 768-dimensional sentence vector from CLS token to one-hot encoded sentence rhetorical roles. The idea is that if certain rhetorical roles are more important than others while creating summaries, then the model will learn those. We call this model BERTSUM RR. Discussion with Legal Experts revealed that ISSUE, RATIO, and RPC are important in summary and must always be selected without the need of summarizing. So we copied all the sentences with predicted rhetorical roles ISSUE, RATIO and RPC regardless of whether they are present in the top 20\% sentences. Model performance evaluated using ROUGE scores \cite{lin2004rouge} are compared in Table \ref{table:ExtractiveSummarizationResults}. Results indicate that rhetorical roles are useful in selecting better summary sentences.

\begin{table}[t]
\small
\begin{center}
\begin{tabularx}{\columnwidth}{|l|X|X|X|}
    \hline
        Model & ROUGE-1 & ROUGE-2 & ROUGE-L \\ \hline
        BERTSUM & 0.60 & 0.42 & 0.59\\ \hline
        BERTSUM RR & 0.62 & 0.46 & 0.61\\ \hline
    \end{tabularx}
\caption{Extractive Summarization Results}
\label{table:ExtractiveSummarizationResults}
 \end{center}
  \vspace{-7mm}
\end{table}

\subsection{Abstractive Summarization of Court Judgments using Rhetorical Roles}

The task of abstractive summarization requires generating concise text summaries of legal documents. For our experiments, we considered 50 randomly selected documents from the Law Briefs dataset (as described in \ref{subsection:extractive_summarization}) as test data. For this task we used pre-trained Legal Pegasus model.\footnote{\url{https://huggingface.co/nsi319/legal-pegasus}} Legal Pegasus is fine-tuned version of Pegasus \cite{zhang2020pegasus} on US securities litigation dataset.\footnote{\url{https://www.sec.gov/litigation/litreleases.htm}} We used the pre-trained Legal Pegasus model for generating abstractive summaries for the baseline. In particular, we split the document into non-overlapping chunks of 1024 tokens, and each chunk was passed through the model to generate summaries. The final summary was obtained by concatenating summaries of each chunk. It constituted the baseline model. We wanted to see how RR could help generate better summaries. Towards this goal, we segmented the document in terms of rhetorical roles, and each of the segments was passed separately through the Legal Pegasus model to generate summaries. The final summary was obtained by concatenating the summaries corresponding to each of the rhetorical roles in the order they appear in the document. This corresponds to the Legal Pegasus RR  model. Both models are compared on the test set and ROUGE scores for both the model are shown in Table \ref{table:AbstractiveSummarizationResults}. As can be observed in Table \ref{table:AbstractiveSummarizationResults} use of rhetorical roles helps to improve the performance on the task of abstractive summarizing. 

\begin{table}[ht]
\small
\begin{center}
\begin{tabularx}{\columnwidth}{|l|X|X|X|}

    \hline
        Model & ROUGE-1 & ROUGE-2 & ROUGE-L \\ \hline
                Legal Pegasus & 0.55 & 0.34 & 0.47 \\ \hline
        Legal Pegasus RR & 0.56 & 0.36 & 0.48 \\ \hline

    \end{tabularx}
\caption{Abstractive Summarization Results}
\label{table:AbstractiveSummarizationResults}
 \vspace{-5mm}
 \end{center}
\end{table}

\subsection{Court Judgment Prediction using Rhetorical Roles}
\newcite{malik-etal-2021-ildc} created the corpus (ILDC: Indian Legal Documents Corpus) and the task (CJPE: Court Judgment Prediction and Explanation) for predicting and explaining the court judgments based on legal judgment texts. It is essential for the judgment prediction task to identify which sentences provide hints about the final decision and use that filtered data as input for prediction. We predicted rhetorical role for each sentence of the train, test data using the baseline rhetorical role model. In the ILDC dataset, we removed the sentences with RPC and RATIO tags making the task more challenging. We also removed the judgments for which no ANALYSIS was predicted. Note that the ILDC dataset is already anonymized and takes care of the biases and ethical concerns associated with the task of judgment prediction. Moreover, we use judgment prediction only as a use case and do not believe that an automated system could remove a human judge; rather, such a system could augment a human and expedite legal processes, especially in highly populated countries like India. 

For the task of judgment prediction, training data had 5044 judgments, and test data had 977 judgments. The idea is to filter the training data using rhetorical roles to check the impact on model performance, keeping the model architecture the same. We used XLNet on the ILDC single model proposed in \newcite{malik-etal-2021-ildc} to predict the judgment outcome on the last 512 tokens of the judgment text. We call this approach XLNet\_last512. The model ran for 13 epochs, and then it was early stopped. In another experiment, we trained the same architecture to predict judgment outcome on the last 512 tokens of ANALYSIS role sentences. We call this model as XLNet\_last512\_Analysis. The model ran for 12 epochs, and then it was early stopped. The model performance comparison are given in Table \ref{table:judgmentPrediction}. As observed from the results, filtering the input text for the ANALYSIS role improves the prediction.

\begin{table}[ht]
\small
\begin{center}
\begin{tabularx}{\columnwidth}{|X|c|c|c|}
    \hline
        \textbf{Model} & \textbf{Precision} & \textbf{Recall} & \textbf{F1} \\ \hline
        XLNet\_last512 & 0.76 & 0.49 & 0.59 \\ \hline
        XLNet\_last512\_Analysis & 0.71 & 0.55 & 0.62 \\ \hline
    \end{tabularx}
\caption{Judgment prediction Results}
\label{table:judgmentPrediction}
 \vspace{-5mm}
 \end{center}
\end{table}

\section{Conclusion and Future Directions}

In this paper, we proposed a new corpus of legal judgment documents annotated with 13 different Rhetorical Roles. The corpus was created via crowdsourcing involving law students. We also proposed baseline models for automatic rhetorical role prediction in a legal document. For some of the roles, the model shows similar trends in predicting the roles as human annotators. Nevertheless, there is scope for further improvement and we have created a leaderboard for the task, so that researchers from community can contribute towards improving the RR prediction system. We also showed two applications of rhetorical roles: summarization and judgment prediction. For both the use-cases use of rhetorical role helps to improve results. We have released the corpus and the baseline models and encourage the community to use these to develop other legal applications as well. 


\section*{Acknowledgements}
We thank EkStep Foundation for funding this work. We thank all the law experts, student volunteers, and coordinators for contributing to data annotation. We thank LawBriefs for sharing the summaries. The author Ashutosh Modi would like to acknowledge the support of Google Research India via the Faculty Research Award Grant 2021.  

\section{Bibliographical References}\label{reference}

\bibliographystyle{lrec2022-bib}
\bibliography{lrec2022-example}

\label{lr:ref}
\bibliographystylelanguageresource{lrec2022-bib}

\end{document}